\documentclass[letterpaper, 10 pt, conference]{ieeeconf}
\IEEEoverridecommandlockouts

\IEEEoverridecommandlockouts                            
\usepackage{cite}
\usepackage{amsmath,amssymb,amsfonts}
\usepackage{algorithmic}
\usepackage{graphicx}
\usepackage{textcomp}
\usepackage{xcolor}
\usepackage[utf8]{inputenc}
\usepackage{graphicx}
\usepackage{tabu}
\usepackage{booktabs}
\usepackage{gensymb}
\usepackage{float}
\restylefloat{table}

\def\BibTeX{{\rm B\kern-.05em{\sc i\kern-.025em b}\kern-.08em
    T\kern-.1667em\lower.7ex\hbox{E}\kern-.125emX}}
\begin{document}

\title{\LARGE \bf
On the Effectiveness of Virtual Reality-based Training for Robotic Setup}

\author{Arian Mehrfard$^{*, 1}$, Javad Fotouhi$^{*, 1}$, \\Tess Forster$^{2}$, Giacomo Taylor$^{2}$, Danyal Fer$^{2}$, \\Deborah Nagle$^{3}$, Nassir Navab$^{1}$, and Bernhard Fuerst$^{2}$ \\ $^{*}$ joint first authors
\thanks{$^{*}$A. Mehrfard and J. Fotouhi are regarded as joint first authors.}
\thanks{This work was partially supported by Verb Surgical Inc.}
\thanks{$^{1}$A. Mehrfard, J. Fotouhi, N. Navab, etc. are with the Laboratory for Computational Sensing and Robotics, The Johns Hopkins University, Baltimore, MD 21218, USA. Corresponding author is Arian Mehrfard at
        {\tt\small amehrfa1@jhu.edu}}%
\thanks{$^{2}$T. Forster, G. Taylor, D. Fer, and B. Fuerst are with Verb Surgical Inc., Mountain View, CA 94043, USA.}%
\thanks{$^{3}$D. Nagle was with Ethicon Inc., Cincinnati, OH 45242, USA}%
}

\maketitle
\thispagestyle{empty}
\pagestyle{empty}

\begin{abstract}
Virtual Reality (VR) is rapidly increasing in popularity as a teaching tool. It allows for the creation of a highly immersive, three-dimensional virtual environment intended to simulate real-life environments. With more robots saturating the industry - from manufacturing to healthcare, there is a need to train end-users on how to set up, operate, tear down, and troubleshoot the robot. Even though VR has become widely used in training surgeons on the psychomotor skills associated with operating the robot, little research has been done to see how the benefits of VR could translate to teaching the bedside staff, tasked with supporting the robot during the full end-to-end surgical procedure. 
We trained 30 participants on how to set up a robotic arm in an environment mimicking clinical setup. We divided these participants equally into 3 groups with one group trained with paper-based instructions, one with video-based instructions and one with VR-based instructions. We then compared and contrasted these three different training methods. 
VR and paper-based were highly favored training mediums over video-based. VR-trained participants achieved slightly higher fidelity of individual robotic joint angles, suggesting better comprehension of the spatial awareness skills necessary to achieve desired arm positioning. In addition, VR resulted in higher reproducibility of setup fidelity and more consistency in user confidence levels as compared to paper and video-based training.
\end{abstract}

\section{Introduction}
Robotics has majorly impacted minimally-invasive surgery. The introduction of this machine into the everyday Operating Room (OR) requires not just the surgeon to be competently trained, but also all the OR support staff. The OR staff is typically trained on how to set up, operate, troubleshoot, and tear down the robotic system.

The current methods for robotic training include a combination of online (e-learning) and instructor-led hands-on training conducted by the manufacturer. Topics covered include system basics, draping the robotic system, patient-focused instructions, and port placement~\cite{green2018current}.

Several studies investigated the effectiveness of Virtual Reality (VR) training for psychomotor skills in both laparoscopic and robotic surgery~\cite{sethi2009validation}. Proficiency-based VR is known to increase the skill level for novice laparoscopic surgeons ~\cite{larsen2009effect,xin2019efficacy}. VR to OR laparoscopic skill transfer has also been studied and validated ~\cite{seymour2002virtual,grantcharov2004randomized,lehmann2005prospective}. In regards to surgical robotic skills, VR simulators improve technical psychomotor performance~\cite{Lerner2010roboticsim}; these technical skills, however, are only applicable for the surgeon. Pertaining to the OR staff, there is a relative lack of research identifying the effects of VR training on the non-technical perioperative skills such as system set-up, draping, and docking.

Recent literature has investigated the use of Augmented Reality (AR) for various surgical robotic tasks~\cite{qian2019review}. Though the majority of the recent research is focused on employing AR as an advanced visualization medium to fuse multi-modal data and promote situation awareness, few works have also discussed the applicability of AR for intra-operative guidance during robot-assisted surgeries. AR was used to project skin landmarks and provide overlay for port placement~\cite{coste2004optimal,falk2005cardio,simoes2013leonardo}. Qian et al. proposed \textit{ARssist} to enhance the hand-eye coordination of the patient-side assistant for tool placement and instrument exchange tasks~\cite{qian2018arssist}. \textit{ARssist} is enabled by rendering the viewing frustum of the laparoscopic camera inside the patient's abdomen. Fotouhi et al. suggested AR support for setting up the robotic arms, assuming a desired joint configuration is known based on the robot's kinemtaics~\cite{fotouhi2020refiective}.

Surgical robotic systems have a steep learning curve for draping, manual alignment of robotic arms to the patient ports, and docking~\cite{iranmanesh2010setup}. These tasks can be a significant factor affecting the overall operation time and experience~\cite{carlsen2014current}. It has been established that the poor placement of ports and robotic components leads to a frustrating experience~\cite{hemal2004nuances}, which may result in collisions, singularities, and an increase in operation time~\cite{liu2019reasons}. 

Current training programs still rely heavily on paper-based instructions and e-learning modules. Given these training platforms, the uncertainty and the lack of standardization around the patient positioning relative to the robotic system remain largely unaddressed. To characterize the robotic staff performances given different training tools, an agnostic robotic arm can be used as a surrogate for a surgical robotic platform. To this end, we used an off-the-shelf seven degree of freedom KUKA iiwa (KUKA AG, Augsburg, Germany). We tasked participants with moving the robotic arm into a predetermined configuration in relation to a manikin. Three training mediums were used: two conventional methods of training, namely paper- and video-based, as well as an immersive VR experience.  


Each participant was asked to complete a training phase, followed by a task execution phase. We compared how each of the training methods impacted the participant's ability to achieve a pre-defined position with optimal robot configuration and patient clearance. Our hypothesis was that VR is an effective, potentially superior, method for teaching the spatial skills associated with setting up a seven degree of freedom robot.

\section{Methods} 
\subsection{Materials}
\begin{figure}[tb]
    \centering
    \includegraphics[width=\columnwidth]{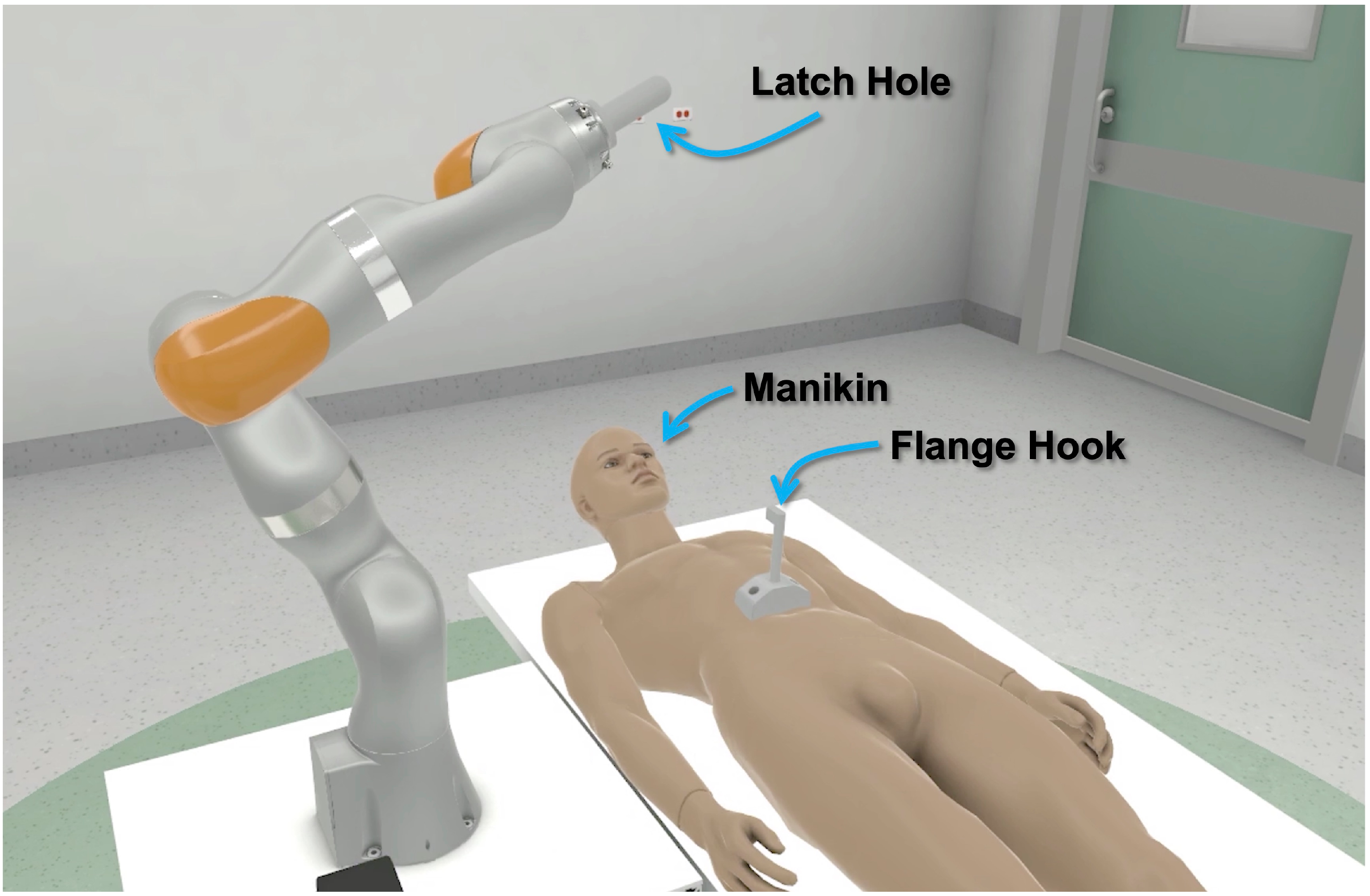}
    \caption{This image shows the initial setup of the experiment, modeled in the virtual world, as each participant finds it at the beginning of the study.}
    \label{fig:setup}
\end{figure}

To keep this study agnostic of surgical robotics supplier, we used a seven degree of freedom industrial robot with seven joints manufactured by the KUKA Robotics Corporation. To simulate the patient in relation to the robotic arm, we used a manikin. As shown in Fig. \ref{fig:setup}, both the manikin and KUKA arm were on separate surfaces. We attached a 3D printed flange to the patient with a male interface while the KUKA arm had a 3D printed female interface on an end-effector. The goal was to dock the flange into the end-effector by following a safe robot trajectory shown in the training phase and avoiding collision with the patient.

\subsection{System Setup}
Video material was shown to the users on a $24"$ display. To deliver the immersive VR training, we used the \textit{HTC Vive Pro} with a dedicated area of $2 \times 3$\,m$^2$ in which the participants could move freely. The VR simulation was executed on an \textit{Alienware (Dell, Round Rock, TX, US)} laptop with an \textit{Intel i7-7700HQ\,k} CPU, \textit{NVIDIA GTX 1070} graphics card, and $16$\,GB RAM.

\begin{figure*}[tb]
     \centering
     \includegraphics[width=\textwidth]{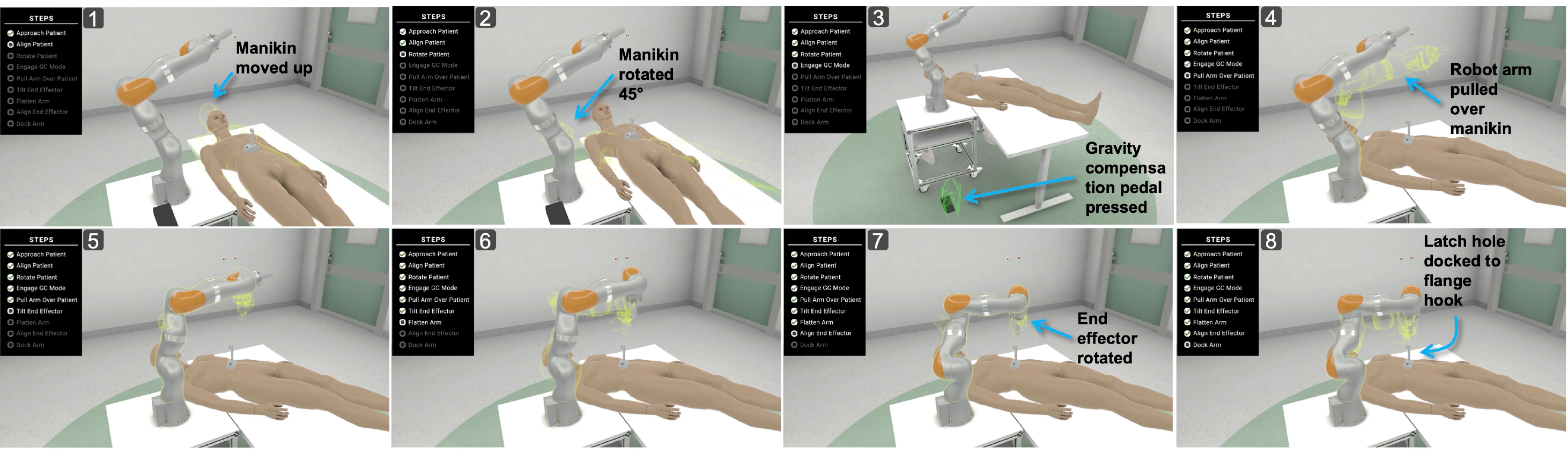}
     \caption{This image shows the initial setup of the experiment, modeled in the virtual world, as each participant finds it at the beginning of the study.}
     \label{fig:stepsFigure}
\end{figure*}

\subsection{Training Approach}
\begin{figure}[tb]
 \centering
 \includegraphics[width=\columnwidth]{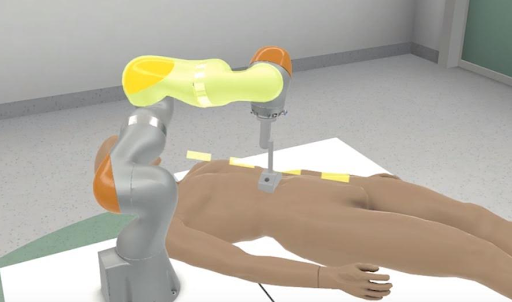}
 \caption{This figure demonstrates the desired end-state of patient and robotic arm. The yellow dotted line depicts the patient center-line, while the highlighted joints are shown as parallel to the patient center-line - ensuring the \textbf{Robot Fidelity Error} is 0$\degree$ in this exemplary state. Note that any deviation from this desired end-state are measured in \textbf{Patient Angle} (45$\degree$ in this state), \textbf{Robot Fidelity Error},  and \textbf{Robot Joint Angles}.}
 \label{fig:endstate}
\end{figure}

The three mediums used for training the participants were paper-based training, video-based training, and VR-based training. To maintain consistency, all three mediums contained content based on the same step-wise approach for setting up the KUKA robot. To this end, the video and paper training materials were derived from the VR application. The VR training included both a passive element of watching the steps play out, with an explanation of the steps, followed by additional guided and unguided steps, comprising the interactive portion of the experience. During the guided component, the users received audio instructions and visual cues on what action they should perform, whereas the unguided component had the users complete the steps without any assistance. The video was a recording of the passive phase of the VR application that was acquired by placing a second virtual camera into the virtual environment. The paper document was based on frame captures from the video and a printed transcript of the audio instructions. The recommended approach was broken down into eight steps, with Fig.~\ref{fig:endstate} depicting the intended end-state of the patient and robotic arm.

Fig.~\ref{fig:stepsFigure} visualizes the following 8 steps, which were included in all three training materials: 
\begin{itemize}
    \item 
    \textbf{Step 1:} Position the patient in relation to a fixed point on the KUKA arm. The participant must position the patient such that the flange $\mathbf{F}$ is in line with the base link $\mathbf{B}$ on the KUKA arm: $\mathbf{F} \times \mathbf{B} = \mathbf{0}$.
\end{itemize}
\begin{itemize}
    \item 
    \textbf{Step 2:} Rotate the patient to 45$\degree$, relative to the table. This is done to ensure the manikin's head is close to the base link of the KUKA arm. 
\end{itemize}
\begin{itemize}
    \item 
    \textbf{Step 3:} Engage gravity-compensation mode, which resists the joint torques, hence counteracts the effects of gravity
\end{itemize}
\begin{itemize}
    \item 
    \textbf{Step 4:} Pull the KUKA arm directly over the patient
\end{itemize}
\begin{itemize}
    \item 
    \textbf{Step 5:} Tilt the end-effector in preparation for docking to the flange 
\end{itemize}
\begin{itemize}
    \item 
    \textbf{Step 6:} Flatten the arm to ensure it is parallel to the manikin's center-line and in position to align and dock the flange to the end-effector.
\end{itemize}
\begin{itemize}
    \item 
    \textbf{Step 7:} Rotate the last robot link and the end-effector simultaneously, to align the male and female connectors for the final docking
\end{itemize}
\begin{itemize}
    \item 
    \textbf{Step 8:} Dock the male connector of the flange into the end-effector
\end{itemize}

During the training phase, each participant absolved precisely one form of training, with a maximum of $20$ minutes to consume the training content. Each participant was required to go through the assigned training content at least twice. The number of times each participant went through the training content, and the total duration of training were recorded. 

This study was reviewed and approved by the Johns Hopkins University Institutional Review Board under the IRB number HIRB00008902. The users were initially screened on whether they are comfortable with wearable technologies. All three groups of subjects were compensated for their voluntary participation. 
    
\subsection{Execution and Metrics}
Regardless of the training method, every participant encountered an execution phase following the training phase. During the execution step, they were asked to perform the steps outlined in training with the KUKA arm, the flange interfaces, and the manikin. It is noteworthy that the point of this study was not to ensure the memorization of the steps during execution, as this is quite rarely required in real-life applications. Therefore, all participants had access to the paper guides, which were displayed on a monitor in the execution room. The performance of the setup, completion of the tasks, and correct alignment of the robot, flange, and patient were evaluated. The metrics and measurements included the following items:
the frequency in which the participants looked at the instructions, angle of the patient to the back of the robot base-link, fidelity of robot arm to patient center-line, and robot joint angles for $6$ relevant joints of the KUKA arm.

\begin{figure}[tb]
 \centering
 \includegraphics[width=\columnwidth]{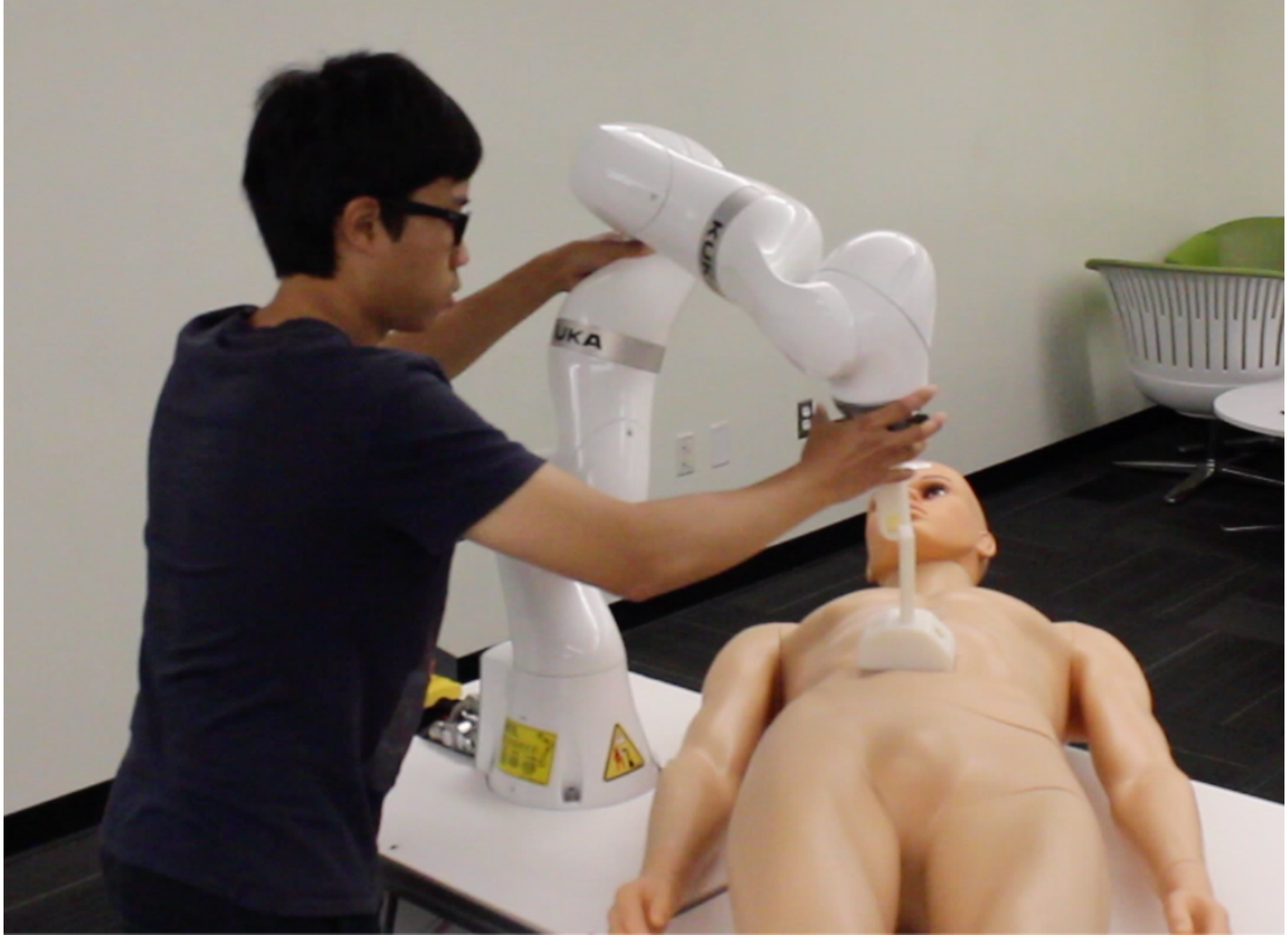}
 \caption{Participant aligns robot with patient after receiving paper training.}
 \label{fig:executionPic}
\end{figure}

\begin{itemize}
    \item The \textbf{Execution Time} was measured from the start until the participant decided that s/he finished the setup: $T = t_\text{end} - t_\text{start}$. The measurements were taken regardless of the completion or correctness of the setup steps.
    \item The order, completion, and correctness of each \textbf{Step} was noted by the proctor. The study execution was not halted in case a step was missed or completed incorrectly. 
    \item The \textbf{Patient Angle Error} describes the angle between the robot base and the midline of the patient model: $\Theta_p = | \theta_\text{robot} - \theta_\text{patient} |$. It was measured by the protractor after each participant completed the setup. In our setup, the desired patient angle was defined as 45$\degree$. 
    \item The \textbf{Robot Fidelity Error} represents the quality of the setup and the misalignment of the desired robot/patient setup. The proctor obtained this error by measuring the angle between the projection of the second to last robot link (shown in yellow in Fig.~\ref{fig:endstate}) onto the patient's surface and the patient center-line: $\Theta_r = | \theta_\text{link} - \theta_\text{patient} |$. Regardless of the patient angle, the optimal robot/patient alignment can be achieved, yielding a robot fidelity error of zero. A perfect robot/patient alignment is illustrated in Fig.~\ref{fig:endstate}.
    \item The six relevant \textbf{Robot Joint Angles} were acquired by interfacing the robot via the MATLAB\textsuperscript{\textregistered} \textit{KUKA Sunrise Toolbox}. The seventh joint angle was ignored as it represents the rotation of the end-effector, which did not affect docking. The difference between the angles in training modules and execution were compared to indicate the correctness of the configuration: $\mathbf{\Theta}_\text{error} = | \mathbf{\theta}_\text{executed} - \mathbf{\theta}_\text{planned} |$, where $\mathbf{\Theta}_\text{error} \in \mathbb{R}^6$.
\end{itemize}

\begin{table*}[t]
  \caption{Mean and standard deviation values for each group. Paper and VR-trained participants spent a similar amount of time training and were highly variable, while the video group consistently spent less time training. While execution time was comparable across all groups, video-trained participants achieved worse alignment in both robot fidelity and patient angle.}
  \label{tab:big_table}
	\centering%
  \begin{tabu} to \textwidth {lllllll}
  \toprule
 & Participants & Training & Execution & Robot Fidelity & Patient Angle & Joint Angle\\
   & & Time (s) & Time (s) & Error ($\degree$) & Error ($\degree$) & Error ($\degree$)\\
  \midrule
    \textbf{Paper} & 10 & 657.8 $\pm$ 333.2 & 235.7 $\pm$  84.3 & 15.0 $\pm$ 13.8 & \textbf{2.6 $\pm$ 2.3} & $39.1\pm55.9$ \\
    \textbf{Video} & 10 & \textbf{484.5 $\pm$ 75.4}  & \textbf{211.3 $\pm$ 128.6} & 25.6 $\pm$ 20.9 & 7.6 $\pm$ 6.3 & $\mathbf{25.4\pm40.4}$ \\
    \textbf{VR}    & 10 & 602.0 $\pm$ 389.5 & 306.3 $\pm$ 137.5 & \textbf{11.4 $\pm$ 16.0} & 5 $\pm$ 3.5 & $28.1\pm51.3$\\
  \bottomrule
  \end{tabu}
\end{table*}

\section{Experiments and Results}
The study population comprised of 30 participants from the students and staff of the Johns Hopkins Whiting School of Engineering. The user subjects had no prior background in performing surgical robot set up. The participants were divided into three equal groups, which determined what training material they were given: paper-based training, video-based training, or VR-based training.

Before the training phase, each participant filled out a questionnaire to ensure his or her aptitude for participation and provided consent to have their initial skill level evaluated.

While training, we measured the time that each participant was engaged with the training material. For the paper-based training group, our measurement associated with the time they read through the document, for the video-based training group we recorded the time they spent watching the material, and for the VR-based training group we started to measure the time from the moment they picked up the Head-Mounted Display (HMD). The time for the VR group included the time it took them to properly mount the HMD on their head until they completed the training application. Even though each participant had to read an instruction document on how to mount the HMD, we did not include this phase in our overall time measurement, as it will become obsolete when people become familiar with the VR HMDs. After the training was completed, the participants scored their own confidence going into the execution phase.

During the execution phase, the execution time, patient angle error, robot fidelity error, robot joint angles, and the correct execution of each step was noted by the proctor. Lastly, a questionnaire was filled out by the participants giving additional feedback and reflecting on their confidence score before the execution.

\begin{table}[]
\centering
\caption{Median values for each training group are presented in the table. Results indicate improved robot alignment for the VR group, and also exhibit a lower median value in ragers to the execution time.}
\label{tab:medians}
\begin{tabular}{llllll}
 & Training & Execution & Fidelity & Patient Angle\\
 & Time (s) & Time (s) & Error ($\degree$) & Error ($\degree$)  \\ \hline
\textbf{Paper} & $582$ & $207$ & $11$ & $\mathbf{2}$  \\
\textbf{Video} & $\mathbf{450}$& $181.5$ & $22$ & $9$ \\
\textbf{VR} & $667.5$ & $\mathbf{155.5}$ & $\mathbf{4.5}$ & $40$ 
\end{tabular}
\end{table}

The means and standard deviations (STD) are reported in Tab.~\ref{tab:big_table}. The average time spent in the training phase of the study for each group was: $658$\,sec for paper (STD: $333$\,sec), $484$\,sec for video (STD: 75\,sec), and $602$\,sec (STD: $390$\,sec) for VR. The average times recorded for the execution phase were: $235$\,sec (STD: $84$\,sec) for paper, $211$\,sec for video (STD: $129$\,sec), and $306$\,sec (STD: $138$\,sec) for VR.

\begin{table}[]
\centering
\caption{The P-values are reported from a one-way ANOVA test on the three participant groups. While there is a trend visible for the VR group to have a better alignment of the robot, it is revealed that the findings are not statistically significant.}
\label{tab:pValues}
\begin{tabular}{llllll}
\textbf{} & Training & Execution & Robot & Patient & Joint \\
 & Time & Time & Error & Angle & Angles \\ \hline
\textbf{P-} & $0.097$ & $0.99$ & $0.19$ & $0.051$ & $0.15$\\
\textbf{value} & & & & &
\end{tabular}
\end{table}

We set the target patient angle to 45$\degree$ in relation to the base. In the evaluation of the patient angle, the paper group performed best with an average error of $2.6\degree$, with VR next at $5\degree$, and video coming in worst with an average error of $7.6\degree$.

\begin{figure}
    \centering
    \includegraphics[width=\columnwidth]{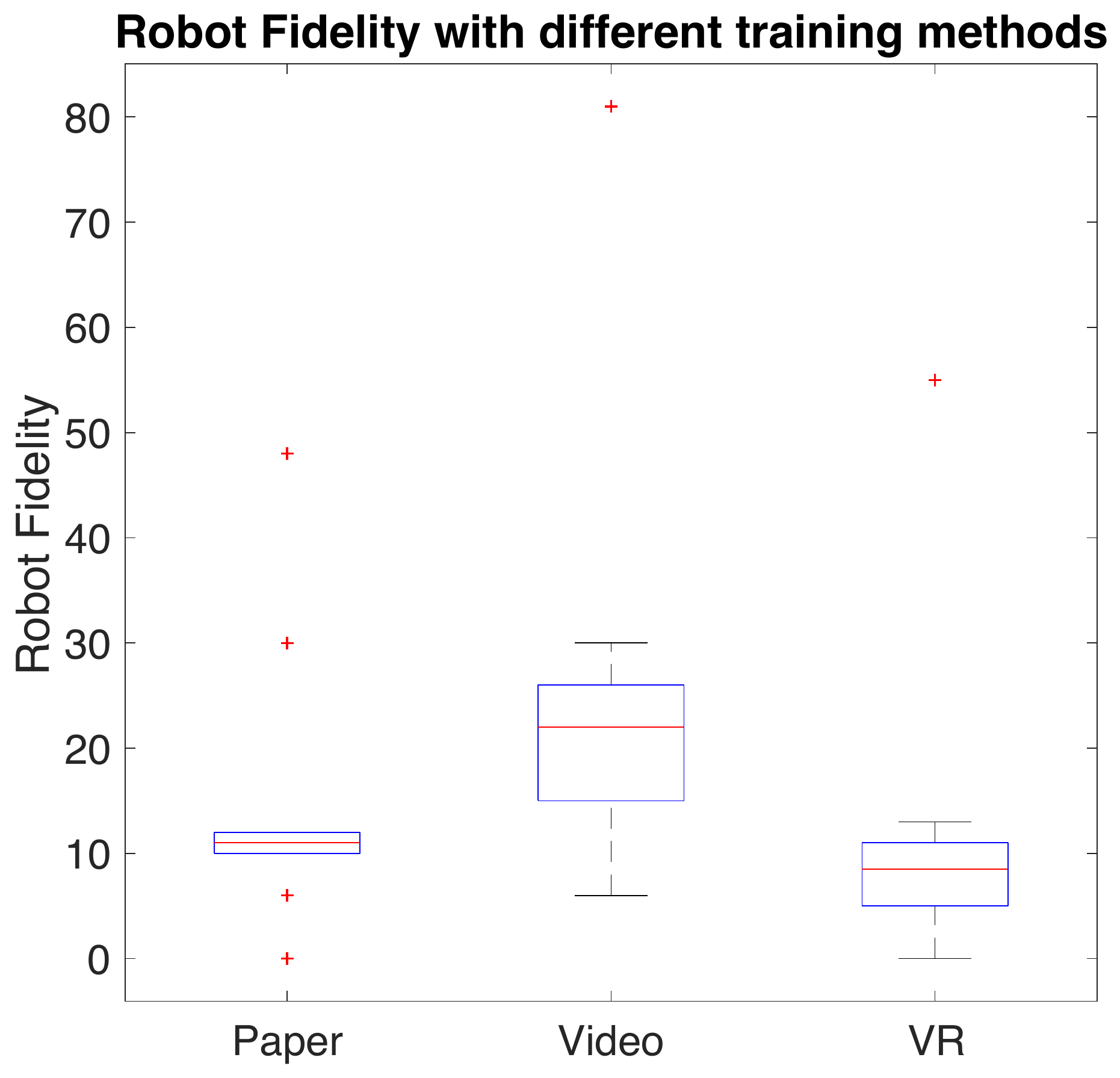}
    \caption{Fidelity error of robotic arm to patient center-line (see Fig.~\ref{fig:endstate} for desired end-state); A participant who achieves highest adherence to this end-state would achieve a robot fidelity error of 0.}
    \label{fig:fidelity}
\end{figure}

The robot fidelity error represents the quality of alignment between the robot orientation and the patient center-line, as depicted in Fig.~\ref{fig:endstate}. Fig.~\ref{fig:fidelity} summarizes the fidelity errors to the pre-defined end-state for each participants' final execution result grouped by the training method. On the whole, participants with VR-based training performed slightly better than the paper-based training, with the video-based group achieving the worst outcome.

\begin{figure}[tb]
    \centering
    \includegraphics[width=\columnwidth]{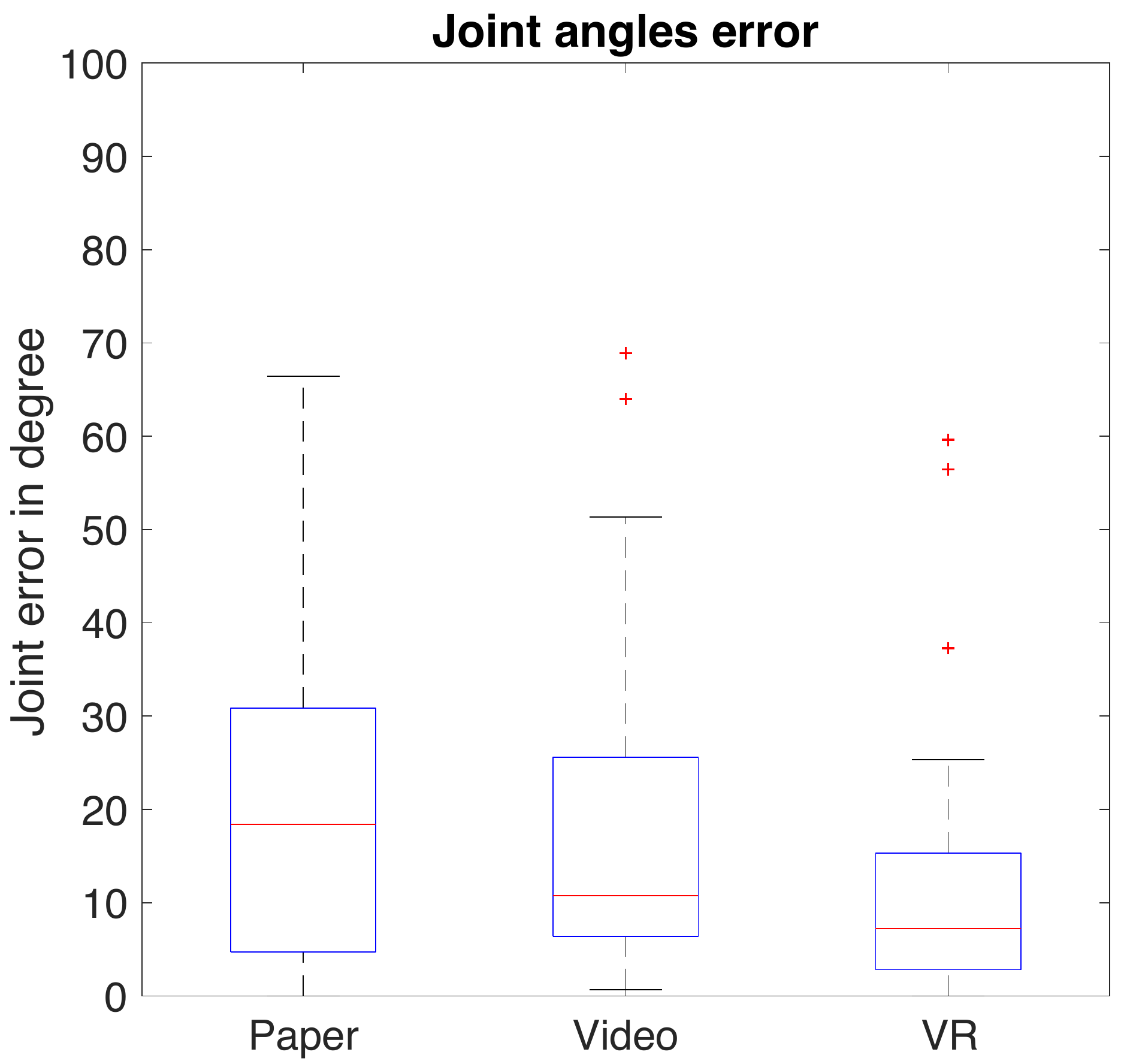}
    \caption{This graph shows the error in each of the six relevant KUKA joint angles for the three groups. These angles were recorded at the end of the execution phase for each participant and compared to the desired end-state.}
    \label{fig:jointErrors}
\end{figure}

\begin{figure}[tb]
    \centering
    \includegraphics[width=\columnwidth]{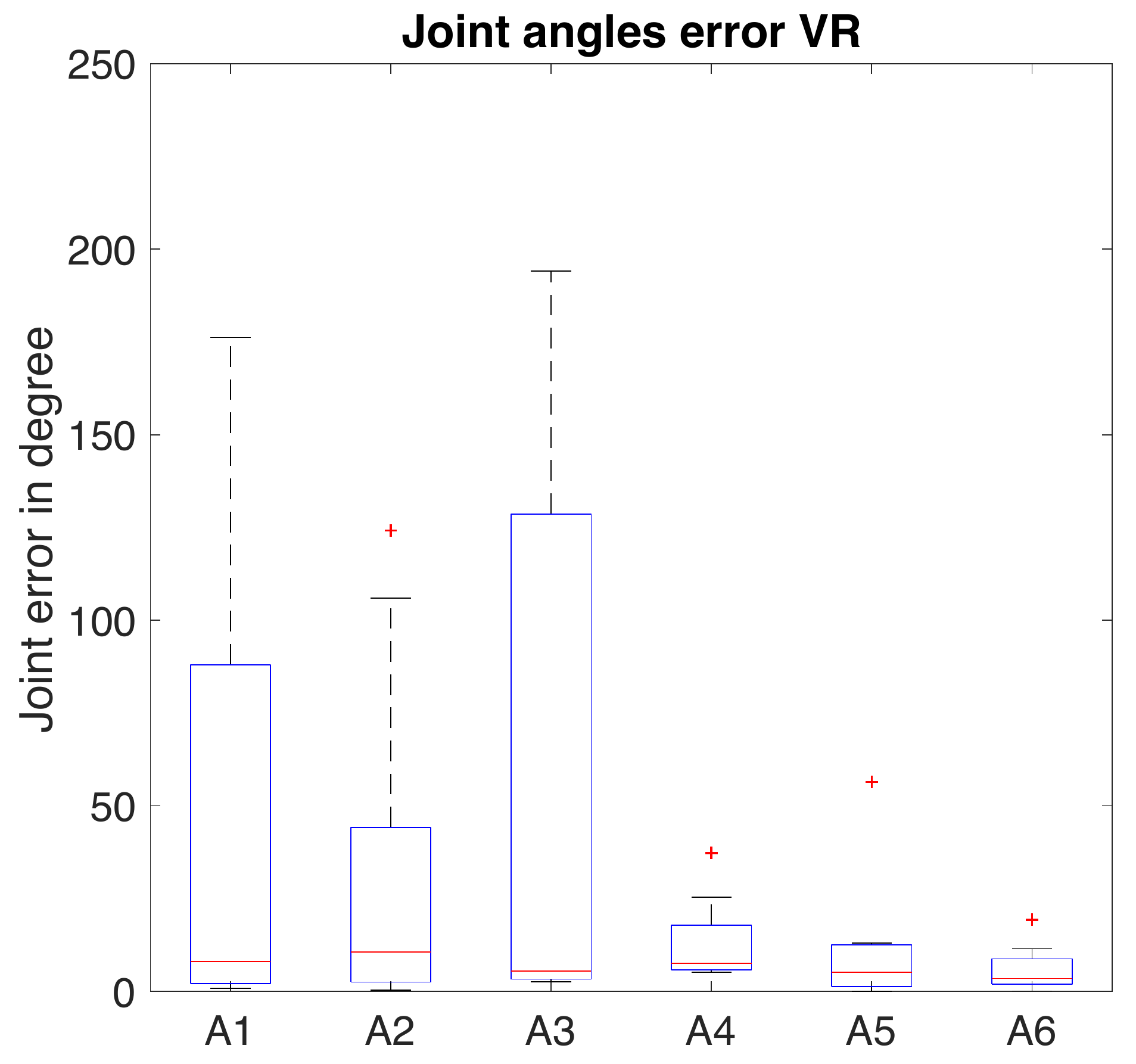}
    \caption{The robot joint angle errors of the VR group. The 7th joint is omitted, as it is merely the rotation of the end effector.}
    \label{fig:jointsVR}
\end{figure}

\begin{figure}[tb]
    \centering
    \includegraphics[width=\columnwidth]{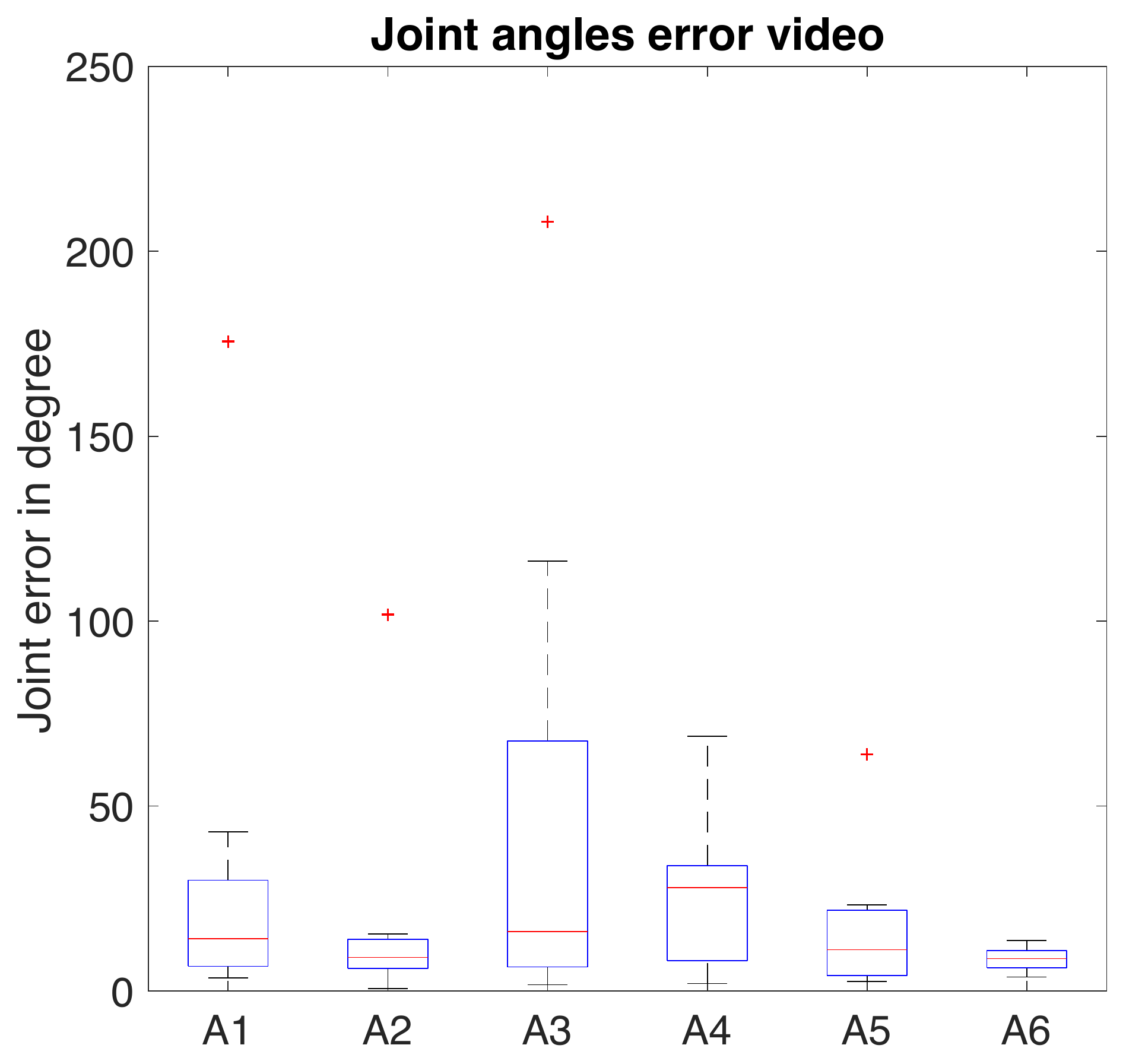}
    \caption{The robot joint angle errors of the video group. The 7th joint is omitted, as it is merely the rotation of the end effector.}
    \label{fig:jointsVideo}
\end{figure}

\begin{figure}[tb]
    \centering
    \includegraphics[width=\columnwidth]{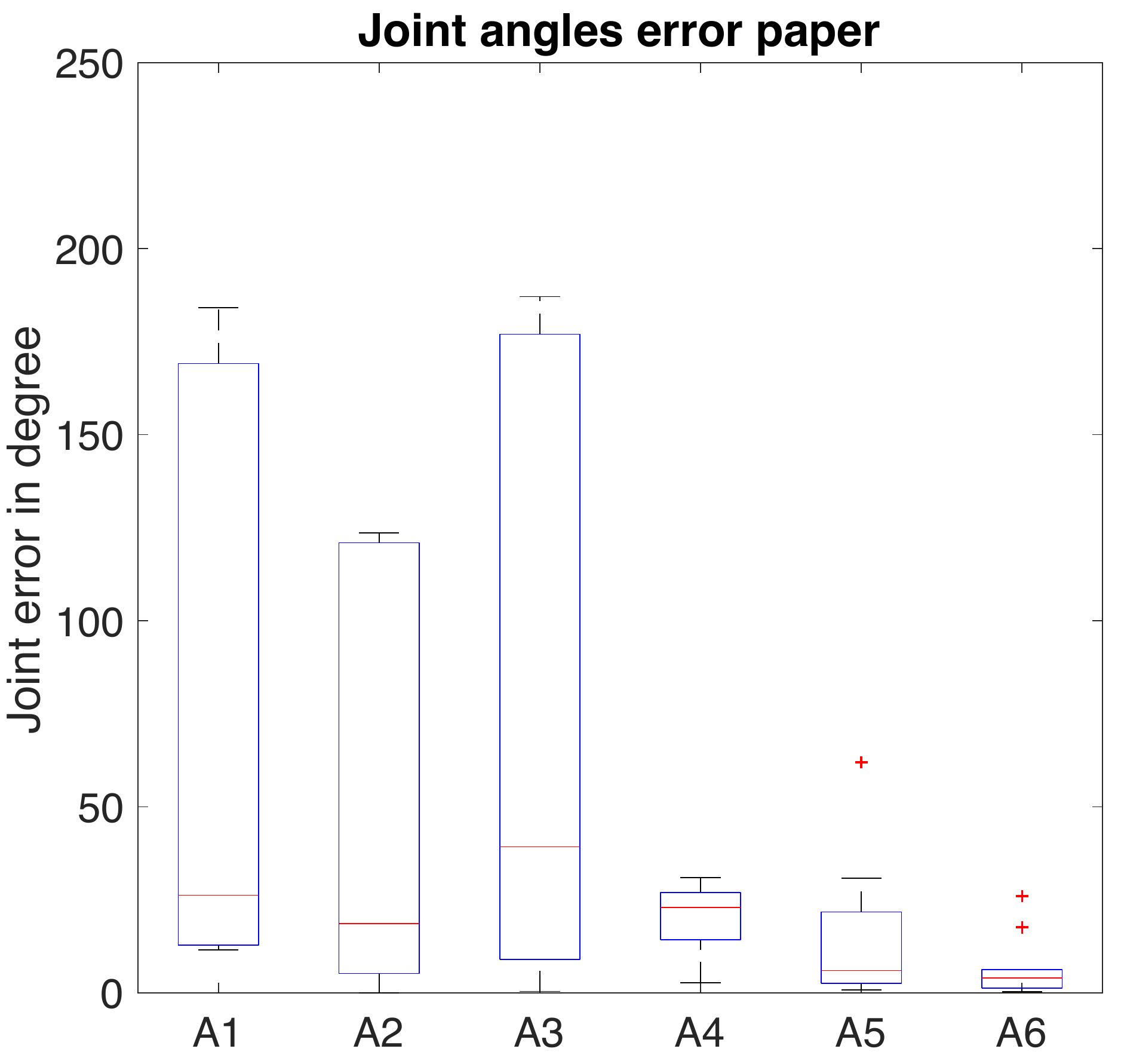}
    \caption{The robot joint angle errors of the paper group. The 7th joint is omitted, as it is merely the rotation of the end effector.}
    \label{fig:jointsPaper}
\end{figure}

The plot in Fig.~\ref{fig:jointErrors} illustrates the errors between the robot joint angles in the training material and the executed joint angles for \hbox{paper-,} video-, and VR-based training, respectively. Additionally Fig.~\ref{fig:jointsVR}-\ref{fig:jointsPaper} give a comparison of each joint angle, for each training group separately. There is a trend towards VR-trained participants performing better than the other two groups. Tab.~\ref{tab:big_table} details further results.

In addition to the mean values, Tab.~\ref{tab:medians} presents the median values for each category. For the training time, the video group had the lowest median of $450$\,sec, which includes the time it took the participants to watch the training video twice. Regarding the execution time and robot fidelity error, the VR group had the lowest median of respectively $155.5$\,sec, and $8.5\degree$. Tab.~\ref{tab:pValues} lists the p-values from a one-way analysis of variance (ANOVA) for the measured metrics, between the three training groups. With this ANOVA we compare the three groups VR, paper and video, and determine if there is a statistically significant difference between their means. Training time has a p-value of $0.097$, execution time has $0.99$, robot fidelity error $0.19$, and patient angle error a p-value of $0.051$, neither of which are statistically significant.

\begin{figure}[tb]
    \centering
    \includegraphics[width=\columnwidth]{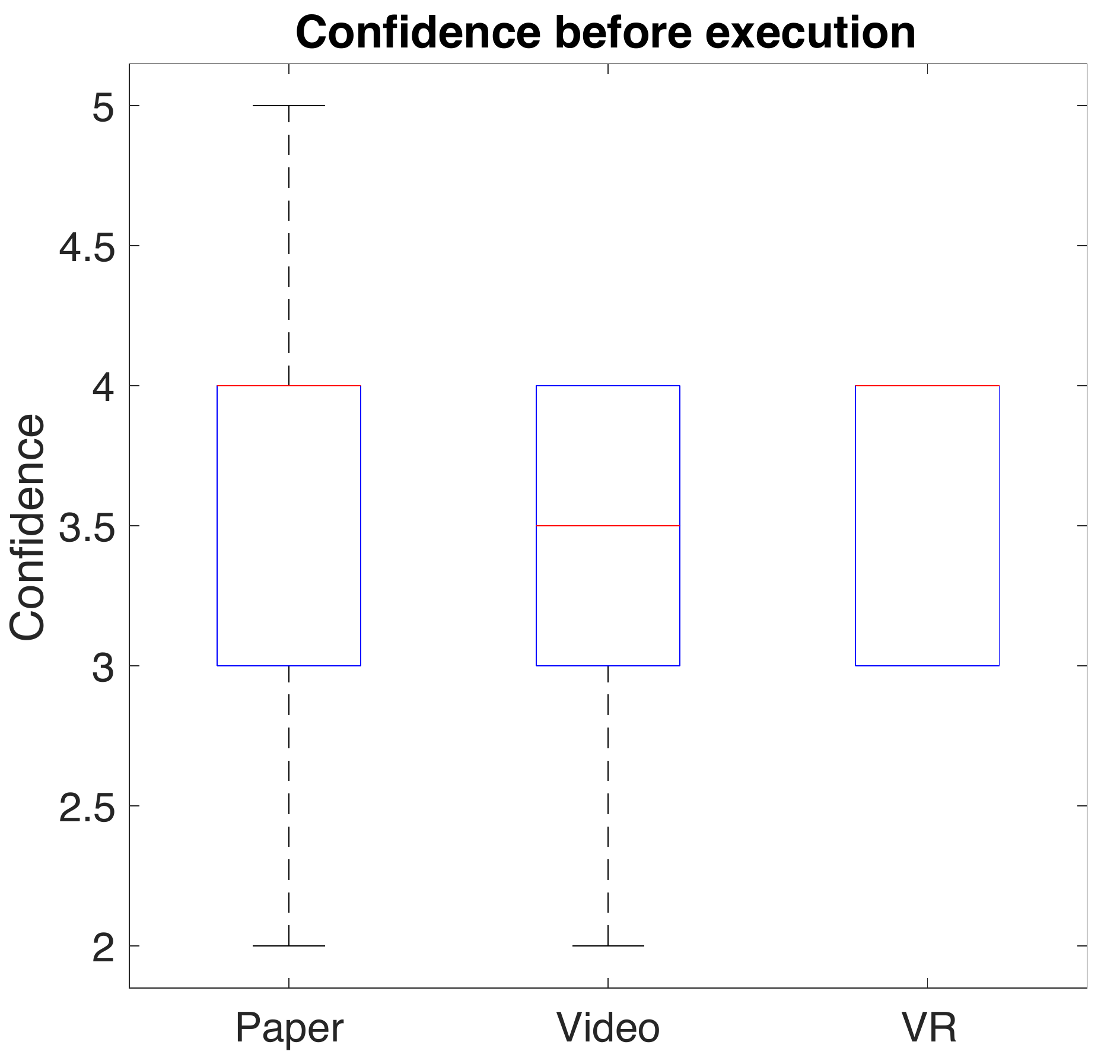}
    \caption{Confidence Rating on a scale of 1-5. 1 represents the lowest confidence and 5 represents the highest. We asked participants to reflect on the accuracy of the confidence score they gave themselves before the execution phase.}
    \label{fig:confidence}
\end{figure}

\begin{figure}[tb]
    \centering
    \includegraphics[width=\columnwidth]{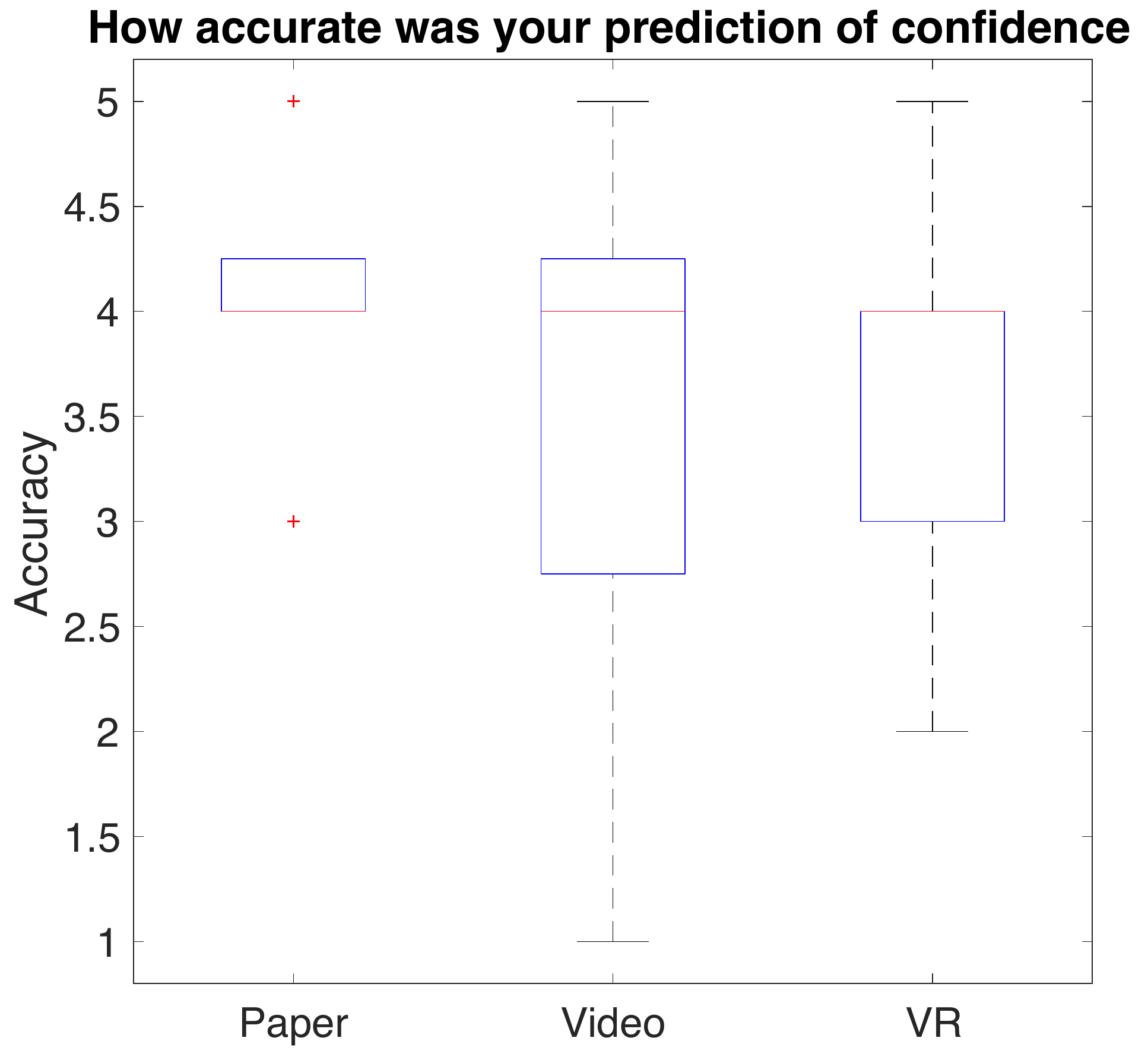}
    \caption{Self reflective accuracy rating on a scale of 1-5. 1 represents the lowest confidence and 5 represents the highest. We asked participants to rank their initial confidence scores after the execution phase.}
    \label{fig:confidenceAfter}
\end{figure}

Fig.~\ref{fig:confidence} shows the confidence score that each participant gave themselves after conduction the training, but before starting the execution phase and Fig.~\ref{fig:confidenceAfter} is based on the self-reflective accuracy score on that confidence, that each user gave themselves after the execution phase. 

\section{Discussion}
\subsection{Training Time}
The video material had a set length of $3$ minutes and $30$\,sec. Participants were asked to watch the entire video, and only two users watched it more than the required two times. In the VR application, users took between $5$ to $8$\,min for the guided and $1.50$ to $4$ minutes for the unguided run. Both these media dictated the pace in which they were consumed, with VR giving the user the option to take longer. In regards to the paper instructions, however, the participants decided on the pace at which they read and comprehended the instructions. This explains the high spread that is visible in the time for paper training. Ignoring the two outliers, the time for VR training is generally narrower and on the lower end compared to paper. Video has the lowest training time; however, as depicted in the figure, it also exhibits the worst results among all three groups.

\subsection{Execution Time}
During the execution time, all users had access to the paper instructions. It is worth considering that the group of participants who trained with the paper instructions took less time to familiarize themselves with the paper instructions in front of them during the execution phase. While VR yielded the highest mean, it also had the highest STD and lowest median (VR=$155.5$\,sec, Video=$181.5$\,sec, Paper=$213$\,sec) which suggests that after disregarding extreme outliers, the VR training had a positive effect on execution time; however, it is not statistically significant (p=$0.99$).

\subsection{Patient Angle}
All training methods instructed the participants to align the patient at a $45\degree$ angle. The paper group consistently came closest to this goal (Paper=$2.6\degree$, Video=$7.6\degree$, VR=$5\degree$). Participants with VR training performed well, particularly when compared to the video group. It should to be noted that these values were not statistically significant (p=$0.051$).

\subsection{Fidelity of Robot Arm to Patient Center-Line}
The end objective presented to the participants was to align the robotic arm to the patient's center-line. The measurement of the fidelity of the robot arm to the patient's center-line, therefore, shows how close the participants achieved this goal. A smaller angle indicates a more accurate alignment and a greater understanding of the end goal. 

The user group receiving VR training had a consistently lower angle, as indicated by both the mean (VR=$11.4\degree$, Video=$25.6\degree$, Paper=$15\degree$) and the median (VR=$8.5\degree$, Video=$22\degree$, Paper=$11\degree$) values. This suggests that VR training leads to a drastically better alignment of the robot arm and, therefore, a higher adherence to the end goal of the task. However, these findings were not statistically significant (p=$0.19$).

\subsection{Fidelity Error of Robotic Arm Joint Angles}
As presented in Fig.~\ref{fig:jointErrors}, after disregarding the outliers, the VR group had the most consistently accurate alignment of the robot joints, supported by the median values (VR=$7.3\degree$, Video=$10.7\degree$, Paper=$18.4\degree$). The mean values (VR=$28.1\degree$, Video=$25.3\degree$, Paper=$39.1\degree$) also show a good performance of the VR group. However, it is surprising that the paper group performed much worse, while the video group has, contrary to the previous data, a proper alignment. One reason for this may be that both VR and the video training depict the movement required by the participant in each step, that is, what motion is required to get the objects (either manikin or robot) from the starting position indicated in each step to the final position of each step. Figs.~\ref{fig:jointsVR}-\ref{fig:jointsPaper} provide a detailed joint-specific representation of the joint angle errors for the VR, video, and the paper groups. They reveal that the joints A1 - A3 were the most susceptible to deficient alignment. It has to be considered that an error of $\sim180\degree$ in these joints suggests that the user rotated the joint in the wrong direction, implying that the symmetry in the links mislead the participants.

\subsection{User Questionnaire}
In Fig.~\ref{fig:confidence}, the confidence ratings show that the VR training group were more consistently confident after training. Fig.~\ref{fig:confidenceAfter} suggests that these participants considered their prediction to be mostly accurate. These claims are supported by free-text remarks that participants made after the execution phase. Most remarks from the paper group revolved around ambiguity and vagueness in the instructions, where participants report to have had trouble understanding specific terms or poses; suggestions were to include a visualization of the robot or to see a video of it moving. From the video group, the users noted the desire for "more visuals", "close up of different angles", and hands-on experience, which are all made available or approximated through the VR application. Participants of the VR group mostly remarked about the inverse kinematics (IK); they noted that the restricted IK simplified the robot and led to unexpected behavior of the real robot. We believe that the full implementation of the IK would allow for an improved skill transfer from the VR application to the real robot. 
 
\subsection{Study Limitations}
The first limitation of this study is related to the implementation of the kinematics for the virtual KUKA robot. To move the arm between states, system-defined touchpoints anchored to the virtual robot model, were grasped by the participant using the VR controllers, and a kinematics backend then solved for joint angles that moved the touchpoint to match the new positioning in real-time. Instead of using a full inverse kinematics model, joints were constrained to only allow for motion along the pre-defined, correct trajectory between waypoints. We can describe this formally, letting $\mathbf{q}\in\mathbb{R}^7$ denoting the joint space configuration of the robot, $\mathbf{p}(\mathbf{q})$ as the forward kinematic cartesian position of the touchpoint, with user-controlled goal position $\mathbf{p}^*$. A full IK model would then move the robot joints in real time to $\mathbf{q}^*$ as found by:
$$\mathbf{q}^*=\underset{\mathbf{q}}{\arg\!\min}\|\mathbf{p}^*-\mathbf{p}(\mathbf{q})\|,$$
subject to joint limits on $\mathbf{q}$. In contrast, our implementation solved for joint configuration $\mathbf{q}^*$ between two known waypoints $\mathbf{q}_1$, $\mathbf{q}_2$ as $\mathbf{q}^*=\mathbf{q}_1+(\mathbf{q}_2-\mathbf{q}_1)t^*$ for
$$t^*=\underset{t}{\arg\!\min}\|\mathbf{p}^*-\mathbf{p}\left(\mathbf{q}_1+(\mathbf{q}_2-\mathbf{q}_1)t\right)\|.$$
The touchpoint was thus constrained to move along a linearly interpolated (in joint space) path between the two waypoints while being dragged. This limitation led to multiple VR-trained subjects to expect the real robot to move similarly between steps. The under-constrained nature of the real robot motion was, therefore, unexpected to a participant and promoted the potential for a negative training effect. A future study should incorporate a full IK model, and possibly incorporate two touchpoints. 

Another limitation that may lead to a negative training effect is the lack of haptic feedback in the VR storyline. The haptic forces a participant would encounter on the actual KUKA robotic arm are noticeable, and yet non-existent in the virtual experience. Some studies have highlighted that haptics inconsistent with real life can lead to a negative effect on trainees~\cite{vaapenstad2017lack,hogle2009validation,westebring2008haptics}. More work needs to be done to determine the extent to which haptics could and should be incorporated into VR training. 

Our participant pool consisted of engineering students and staff who did not have a healthcare background. Those with healthcare backgrounds, particularly the OR staff, are likely to understand the context of the task better and therefore be more diligent in following the directions. For example, an OR staff member might better comprehend the implications of, and therefore better perform, patient positioning (as measured in this study by patient angle). A larger participant pool with more relevant background could provide a more fair and holistic comparison between the training methods.

Last but not least, the crucial challenge in conducting such a study was the new and unfamiliar VR environment for the participants that did not have experience with immersive technologies. Future studies should consider the learning curve and train users in multiple sessions (maybe weeks apart), such that they get comfortable with the study material.

\section{Conclusions}
To the best of our knowledge, this is the first study that systematically compared training technologies for critical tasks performed by OR staff. Although fundamentally different in nature, we have built paper-, video-, and VR-based training materials containing the same information, and observed a surprising difference in training outcome. 

Our training goals were deliberately defined to transfer knowledge of spatial awareness and physical actions. The results and conclusions did not directly translate to training goals that pertain to clinical knowledge or psychomotor skills (e.g., surgical tasks). The deployment of novel technologies, such as VR, for training and education, needs to be carefully evaluated on a case-by-case basis. 

VR- and paper-based training have performed similarly, while video-based training yielded the most inaccurate alignment. The most accurate joint angles were achieved by VR-trained participants, indicating that VR is a suitable tool to train spatial awareness. Furthermore, the self-reported confidence levels prior to task execution were more reproducible with VR training and are likely to be increased through further improvements of a VR experience. Several studies have ascertained that the increased confidence as a result of effective training leads to trainees that are more likely to transfer what they have learned to occupational tasks~\cite{gist1989alternative, tannenbaum1991expectations,stajkovic1998selfefficacy,burke2007trainingtransfer}.

When combining our observations of a more consistent self-reported confidence and the desire of the paper-based learners for videos or interactive material to supplement the paper-based experience, we can conclude that VR represents a more engaging and entertaining experience over traditional training. Ultimately, this is a crucial aspect, noting that continued engagement with training materials tends to yield a better education outcome. 


\nocite{*}
\bibliographystyle{unsrt}
\bibliography{literature}
\end{document}